\documentclass{article}
\usepackage{tikz}
\usepackage{spconf,amsmath,graphicx}
\usepackage{algorithm}
\usepackage{booktabs}
\usepackage{amssymb}
\usepackage{algorithmicx}  
\usepackage[noend]{algpseudocode}
\usepackage{algpseudocode} 
\usepackage{stfloats}

\title{Ladder-of-Thought: Using Knowledge as Steps to Elevate Stance Detection}
\name{Kairui Hu$^{\star}$, Ming Yan$^{\star}$, Joey Tianyi Zhou$^{\star}$, Ivor W. Tsang$^{\star}$, Wen Haw Chong$^{\dagger}$, Yong Keong Yap$^{\dagger}$}

\address{$^{\star}$ Centre for Frontier AI Research (CFAR), A*STAR, Singapore \\
$^{\star}$ Institute of High Performance Computing (IHPC), A*STAR, Singapore \\
  $^{\dagger}$ DSO National Laboratories, Singapore}

\begin{document}
%
\maketitle

\begin{abstract}


Stance detection aims to identify the attitude expressed in a document towards a given target. Techniques such as Chain-of-Thought (CoT) prompting have advanced this task, enhancing a model's reasoning capabilities through the derivation of intermediate rationales. However, CoT relies primarily on a model's pre-trained internal knowledge during reasoning, thereby neglecting the valuable external information that is previously unknown to the model. This omission, especially within the unsupervised reasoning process, can affect the model's overall performance. Moreover, while CoT enhances Large Language Models (LLMs), smaller LMs, though efficient operationally, face challenges in delivering nuanced reasoning.
In response to these identified gaps, we introduce the \textbf{L}adder-\textbf{o}f-\textbf{T}hought (LoT) for the stance detection task. Constructed through a dual-phase Progressive Optimization Framework, LoT directs the small LMs to assimilate high-quality external knowledge, refining the intermediate rationales produced. These bolstered rationales subsequently serve as the foundation for more precise predictions - akin to how a ladder facilitates reaching elevated goals. LoT achieves a balance between efficiency and performance. Our empirical evaluations underscore LoT's efficacy, marking a $16\%$ improvement over GPT-3.5 and a $10\%$ enhancement compared to GPT-3.5 with CoT on stance detection task.

\end{abstract}

\begin{keywords}
Stance Detection, Ladder-of-Thought, Language Model, Knowledge Infusion
\end{keywords}
\section{Introduction}
\label{sec:intro}

Stance detection is the task of discerning the stance towards a specific target in an provided document. This task can be challenging given the breadth of topics and the depth of reasoning required to make accurate predictions.
Nevertheless, the landscape of stance detection has evolved significantly with the success of Pre-trained Language Models (PLMs). These PLMs, when fine-tuned for downstream tasks, demonstrate a remarkable improvement in performance \cite{zhang2023stance, cke-net}.

\begin{table}[ht]
\centering
\begin{tabular}{l@{\hspace{0.17cm}}c@{\hspace{0.17cm}}c@{\hspace{0.17cm}}c@{\hspace{0.17cm}}c}
\toprule
\textbf{Paradigms} & \textbf{Knowledge} & \textbf{Sizes} &\textbf{Reasoning} & \textbf{Performance} \\
\midrule
WS-BERT             & External & 340\textit{M} & Weak   & 74.5 \\
CoT                 & Internal & 175\textit{B} & Strong & 68.9 \\
\textbf{LoT} (ours) & External & 780\textit{M} & Strong & 79.2 \\
\bottomrule
\end{tabular}
\caption{Comparison of different stance detection paradigms.}
\label{table:performance}
\end{table}

Leveraging the capabilities of LMs, prompt-based techniques have further enhanced the performance, especially when LLMs such as GPT-3.5 are equipped with meticulously designed prompts \cite{chatgpt-cot}. The Chain-of-Thought (CoT) prompting stands as a prominent prompting strategy, enabling LMs to produce coherent and systematic reasoning rationales, which in turn improves the subsequent prediction accuracy \cite{cot}. However, CoT has a discernible limitation: it mainly relies on the model's internal, pre-existing knowledge when generating these rationales \cite{verifyandedit}. External knowledge, which is often dynamic, evolving, and abundant in domain-specific insights, remains unexploited \cite{cot-knowledge}. Given CoT's reliance on the model's pre-trained knowledge, its unsupervised intermediate reasoning process may inevitably produce less reliable rationales, affecting the model's overall performance \cite{verifyandedit, cot-knowledge, yang2023chatgpt, unrely-2}.

The integration of external background knowledge is paramount for optimizing models' stance detection capabilities \cite{wsbert}. Predictions can be compromised in the absence of this auxiliary information, particularly when limited by the model's intrinsic knowledge. Table 1 serves as a testament: despite ChatGPT's utilization of CoT \cite{chatgpt-cot}, smaller models like BERT can outperform it in stance detection tasks when supplemented with external knowledge from Wikipedia \cite{wsbert}.

Moreover, the expansive architecture of LLMs like GPT-3.5 brings concerns about efficiency. On the other hand, smaller LMs, though more operationally efficient, often compromise on the reasoning capability due to their compactness \cite{cot, yang2023chatgpt}.
And while CoT provides performance gain in LLMs, it does not effectively benefit the smaller-sized models \cite{cot}. This underscores the need for enhancing the reasoning prowess of smaller models without bloating their size.

\begin{figure*}[ht]
  \centering
  \includegraphics[width=\textwidth]{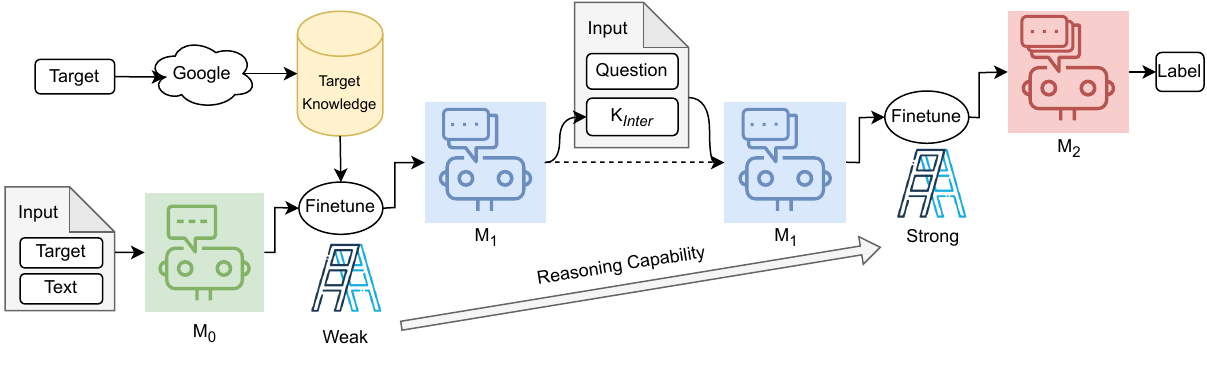}
  \caption{The Overview of Ladder-of-Thought Architecture}
  \label{architecture}
\end{figure*}

To address these challenges, we propose \textbf{L}adder-\textbf{o}f-\textbf{T}hought (LoT), a novel methodology that leverages external knowledge as steps to elevate stance detection. LoT operates on a Progressive Optimization Framework. The "ladder" in LoT represents this Progressive Optimization Process. The initial phase absorbs external information, guiding the model to generate more reliable intermediate knowledge as rationales. This intermediate knowledge, act as "steps" to progressively elevate the model's comprehensive understanding capability, culminating in a robust stance detection. Tailored for smaller LMs, LoT strikes a harmonious balance between efficiency and performance. It facilitates the seamless integration of ample external knowledge and cultivates profound reasoning capabilities. 
The architecture of LoT is illustrated in Fig. \ref{architecture}.

Our main contributions are summarized as follows: 
\begin{itemize}
    \item We introduce LoT – a novel method for stance detection. By enriching smaller LMs with external knowledge, LoT effectively facilitates the generation of more reliable intermediate rationales, consequently enhancing the prediction performance.
    \item We demonstrate that LoT outperforms existing methods, achieving state-of-the-art results while maintaining efficiency.
\end{itemize}


\section{Methodology}
\label{sec:meth}

\subsection{Task Definition:}
\label{ssec:task}
\noindent \textbf{Stance Detection}: Stance detection involves identifying the stance of an opinionated document concerning a specific target. Formally, consider a set \(D = \{(x_i = (d_i, t_i), y_i)\}_{i=1}^n\) representing \(n\) instances. Here, \(x_i\) encapsulates a document \(d_i\) and a target \(t_i\). The task is to deduce the stance label, \(y_i\), which can be categorized as \(\{\text{positive, negative, neutral}\}\). 

\subsection{External Knowledge Retrieval}
\label{ssec:external}
To increase the reliability of the generated intermediate rationales in LoT, we integrate external knowledge to enhance the generation in a supervised manner. Specifically, a web retrieval process fetches pertinent external information for each target \(t_i\) from Google Search. By extending beyond the traditional realms of Wikipedia and diving into the wider web, we access a plethora of diverse and dynamic information \cite{google}. This shift aligns with the emerging trend of exploring beyond the boundaries of Wikipedia-based research \cite{webgpt, piktus2022web, google}.

\subsection{Ladder-of-Thought (LoT) Architecture:}
\label{ssec:lot}
The Ladder-of-Thought (LoT) architecture enhances stance detection, enabling smaller models to reason more effectively. LoT draws its metaphor from the construction of a ladder, where the process of Progressive Optimization forms the framework of the ladder, and the reliable intermediate knowledge, fortified with external insights, serves as the integral "steps". These pivotal steps empower the model to reach heightened insights and deeper comprehension, facilitating more accurate predictions. LoT is developed through a dual-phase Progressive Optimization Framework:

\begin{enumerate}
    \item \textbf{Phase 1 - Generation Fine-tuning}: In this foundational phase, the pre-trained model \(M_0\) is fine-tuned with the retrieved knowledge. This transfers the external insights to the model, guiding it to generate more robust intermediate knowledge that subsequently aids in downstream stance predictions. The resulting model \(M_1\) facilitates the generation of more enriched and reliable intermediate rationales, denoted as \(k_{\textit{intermediate\_i}}\).

    \item \textbf{Phase 2 - Prediction Fine-tuning}: Phase-2 utilizes the enhanced knowledge generated from Phase-1 to expertly discern stance labels. By concatenating the document, target, and the generated knowledge, we construct an enhanced input representation, \(x_{\textit{enhanced\_i}}\). \(M_1\) is then fine-tuned with this enhanced input, culminating in the final model \(M_2\). Given the knowledge-infused input, \(M_2\) can conduct stance prediction \(y_i\). 

\end{enumerate}

\noindent The Ladder-of-Thought (LoT) architecture employs a Progressive Optimization Framework to enhance the stance detection model step-by-step. Leveraging the concept of cognitive evolution, LoT signifies a novel paradigm for model training. In particular, phase-1 is the foundation of LoT, infusing the model with core knowledge, reminiscent of grounding a student in fundamental theories. In Phase-2, this grounded rationale is utilized to guide the model towards more nuanced stance detection. The optimization from \( M_0 \) to \( M_2 \) via \( M_1 \) reflects the LoT philosophy: evolving model capabilities through deliberate optimization, striking a balance between computational efficiency and reasoning depth.

For a detailed step-by-step procedure of the Progressive Optimization, refer to Algorithm~\ref{alg:LoT}.

\floatname{algorithm}{Algorithm}
\renewcommand{\algorithmicrequire}{\textbf{Input: }}
\renewcommand{\algorithmicensure}{\textbf{Output: }}

\begin{algorithm}
    \caption{Progressive Optimization Algorithm}
    \label{alg:LoT}
    \begin{algorithmic}[1]
        \Require Document matrix \( D = \{d_1, d_2, ..., d_n\} \), 
         Target vector \( T = \{t_1, t_2, ..., t_n\} \), Pre-trained model \( M_0 \)
        \Ensure Stance prediction vector \( Y = \{y_1, y_2, ..., y_n\} \)
        
        \Function{LoT}{$D, T, M_0$}
            \State \textbf{Phase-1}: 
            \For{\( i = 1 \) to \( n \)}
                \State \( k_i \leftarrow \text{WebRetrieval}(t_i) \)
            \EndFor

            \State \( M_1 \leftarrow \text{GenerationFinetune}(\{k_1, k_2, ..., k_n\}, M_0) \)
            
            \For{\( i = 1 \) to \( n \)}
                \State \( k_{\textit{intermediate\_i}} \leftarrow M_1(d_i, t_i) \)
                \State \( x_{\textit{enhanced\_i}} \leftarrow \text{IntegrateInputs}(d_i, k_{\textit{intermediate\_i}}, t_i) \)
            \EndFor

            \State \textbf{Phase-2}: 
            \State \( M_2 \leftarrow \text{PredictionFinetune}(\{x_{\textit{enhanced\_1}}, x_{\textit{enhanced\_2}}, ..., \) 
            \Statex \hspace{\algorithmicindent} \(x_{\textit{enhanced\_n}}\}, M_1) \)
            
            \For{\( i = 1 \) to \( n \)}
                \State \( y_i \leftarrow M_2(x_{\textit{enhanced\_i}}) \)
            \EndFor
            
            \State \Return \( Y \)
        \EndFunction  
    \end{algorithmic}
\end{algorithm}

\section{Experiment}
\label{sec:exp}
\subsection{Dataset and Evaluation Metric}
\label{ssec:dataset}

The VAried Stance Topics (VAST) \cite{vast} is a classic zero-shot and few-shot stance detection dataset. It encompasses a broad spectrum of topics: 4,003 for training, 383 for development, and 600 for testing. Unlike other datasets for stance detection like P-stance \cite{p-stance} which only have 2 targets or SemEval-2016 \cite{semeval} with 4 targets, VAST covers a numerous number of targets spanning various domains. Following the preceding studies \cite{vast, wsbert}, the macro average of F1-score is used as the evaluation metric.

\subsection{Baselines and Models}
\label{ssec:model}
We employ FLAN-T5-Large, the 780M parameter version of FLAN-T5, as our backbone. We compare our model with the following baselines: TGA-Net \cite{vast}, BERT, BERT-GCN \cite{bertgcn}, CKE-Net \cite{cke-net}, WS-BERT-Single \cite{wsbert}, DQA \cite{chatgpt-cot}, StSQA \cite{chatgpt-cot}. The first five methods are based on BERT and its variants. DQA is based on ChatGPT with direct input-output (IO) prompting, while StSQA employs CoT on ChatGPT, prompting ChatGPT in a step-by-step manner. 

\subsection{Result}
\label{ssec:result}

The overall results of our model and the baselines are reported in Table \ref{table:performance}.

\begin{table}[ht]
\centering
\begin{tabular}{@{} l l r c @{}}
\toprule
\textbf{Methods} & \textbf{Models} & \textbf{F1 Scores} \\
\midrule
TGA-Net & BERT  & 66.5 \\
BERT & BERT  & 68.4 \\
BERT-GCN & BERT  & 69.2 \\
CKE-Net & BERT & 70.1\\
WS-BERT-Single & BERT  & 74.5 \\
DQA & GPT-3.5  & 62.3 \\
StSQA & GPT-3.5 & 68.9 \\
\midrule
Baseline FLAN-T5 & FLAN-T5  & 73.6 \\
\textbf{LoT} (Ours) & FLAN-T5 & \textbf{79.2} \\
\bottomrule
\end{tabular}
\caption{Performance comparison on the VAST dataset.}
\label{table:performance}
\end{table}

Compared to the baseline FLAN-T5, LoT achieves a remarkable improvement, achieving an F1 score of $79.2$, while FLAN-T5 achieves an F1 score of $73.6$. This highlights the efficacy of our LoT. Furthermore, compared to ChatGPT-based DQA, which operates on an expansive architecture and achieves an F1 score of $62.3$, our LoT demonstrates not just superior performance but tangible efficiency with significantly fewer parameters. This compact model size promises better deployment possibilities in real-world scenarios where computational resources can be a constraint.

Compared to StSQA with an F1 score of $68.9$, our LoT also outperforms this CoT-enhanced ChatGPT approach. This result showcases that despite CoT amplifying internal reasoning, our LoT can absorb high-quality external knowledge, facilitating a more accurate prediction.

\subsection{Ablation Study}
\label{ssec:abl}

The foundational structure of LoT is built on the dual-phase Progressive Optimization framework. As all implementations involves the Prediction Fine-tuning, our focus lies in understanding the efficacy of two specific aspects of LoT: Generation Fine-tuning and the enhanced intermediate knowledge. We conduct an ablation study to evaluate their individual and comprehensive impact. In addition to the baseline and the complete LoT implementation, we introduce two intermediate settings for comprehensive comparison:

\textbf{CoT}: Following the principle of CoT, this configuration skips Generation Fine-tuning, directly utilizing the pre-trained model to produce intermediate knowledge and perform the subsequent prediction. This offers insights into the impact of the raw knowledge that is directly prompted from the pre-trained model on prediction performance.

\textbf{Phase1-Only}: Focusing exclusively on Phase-1 Fine-tuning, this configuration omits the subsequent integration of the generated knowledge during Phase-2 fine-tuning. The objective is to evaluate the direct influence of Phase-1 Fine-tuning and determine if it enhances the model's intrinsic knowledge.

\begin{table}[ht]
\centering
\begin{tabular}{@{} l c c r c @{}}
\toprule
\textbf{Models} & \textbf{Gen Fine-tuning} & \textbf{Gen Knowledge} & \textbf{F1} \\
\midrule
Baseline & -- & -- & 73.4\\
CoT & -- & \checkmark & 73.1\\
Phase1-Only & \checkmark & -- & 74.2\\
LoT & \checkmark & \checkmark & 79.2\\
\bottomrule
\end{tabular}
\caption{Ablation study on LoT.}
\label{table-abl}
\end{table}

Table \ref{table-abl} showcases the results of the ablation study.

The Baseline achieves an F1 score of 73.4, representing the performance without any additional enhancements. 

By comparison, the CoT configuration slightly decreases to 73.1. This aligns with our prior discussion that small models may not benefit from CoT due to their limited reasoning capabilities \cite{cot}. Although directly prompting for intermediate knowledge yields some rationales, their quality is compromised. The unsupervised nature of these intermediate outputs may introduce potential noise. Hence, introducing CoT might inadvertently add complexity to the models, distracting them from accurate prediction. This underscores the significance of a supervised fine-tuning phase to enhance the reliability in knowledge generation.

The Phase1-Only configuration achieves an F1 score of 74.2, surpassing our baseline. This score suggests that Generation Fine-tuning can effectively enhance the model's inherent knowledge base. By supplementing the model with external information, even without explicitly leveraging the generated knowledge during predictions, we can still witness an improvement over the baseline performance. This underscores that enriching the foundational knowledge of the model can inherently bolster its capabilities in stance detection.

With our LoT configuration, the model reaches an F1 score of 79.2, showcasing a remarkable performance improvement over both the baseline and the other configurations. This substantial increase underlines the benefits of our overall Progressive Optimization Framework in LoT.

\subsection{Overfitting in Progressive Optimization}
\label{ssec:overfit}
\begin{figure}[ht]
  \centering
  \includegraphics[width=0.45\textwidth]{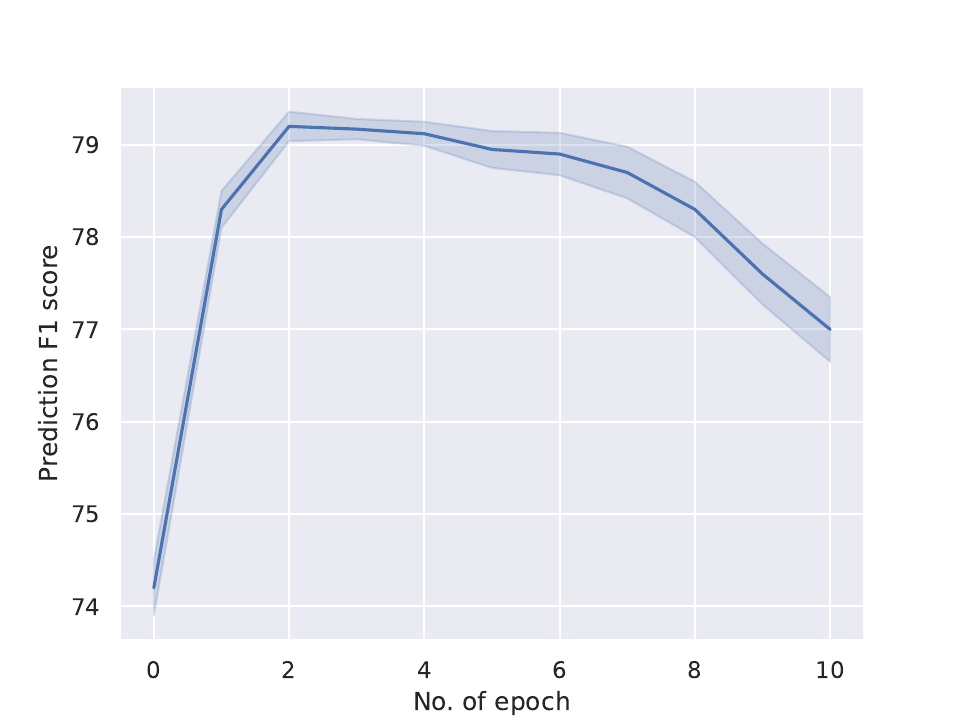}
  \caption{Effect of Phase-1 Training epochs on the overall prediction accuracy.}
  \label{epoch}
\end{figure}

In our Progressive Optimization Framework, overfitting presents a notable challenge. If the model undergoes excessive training during Generation Fine-tuning (Phase-1), it could become overly specialized for the initial task, leading to a detrimental impact on its performance during the subsequent predictions. Achieving the ideal equilibrium between these phases is crucial. We investigate the influence of training epochs in Phase-1 on the subsequent prediction accuracy in Phase-2. The outcomes are depicted in Figure \ref{epoch}.

The findings suggest that the optimal performance is achieved at around 2 epochs, with a subsequent decline in performance as the number of epoch increases. This juncture signifies the ideal balance: it facilitates the generation of high-quality intermediate knowledge without an excessive reliance on Phase-1. While Phase-1 aims to enhance the model's reasoning for Phase-2, it is important to avoid overemphasizing the former phase at the expense of the latter. Our results highlight the importance of a strategic equilibrium, ensuring that each phase complements the other, ultimately constructing a robust and effective Progressive Optimization Framework.

\section{Conclusion}
\label{sec:con}

In this research, we introduce the Ladder-of-thought (LoT). This method effectively enhances the smaller LMs' reasoning abilities with a dual-phase Progressive Optimization Framework. LoT enables the model to efficiently absorb high-quality external knowledge, thereby crafting more reliable intermediate rationales that facilitate accurate predictions. Our empirical evaluations demonstrate the efficacy of LoT, highlighting its superiority over the existing methods. LoT showcases that even smaller LMs, with the right guidance, can outperform LLMs like ChatGPT in stance detection. LoT is also applicable to other downstream tasks, and we aim to explore further in future works. 

\bibliographystyle{IEEEbib}
\bibliography{refs}

\end{document}